\documentclass[letterpaper, 10 pt, conference]{ieeeconf}  % Comment this line out if you need a4paper

\IEEEoverridecommandlockouts                              % This command is only needed if 
\title{\LARGE \bf
Compact LED-Based Displacement Sensing for Robot Fingers}

\author{
Amr El-Azizi\authorrefmark{1}\authorrefmark{2}, 
Sharfin Islam\authorrefmark{1}\authorrefmark{2}, 
Pedro Piacenza\authorrefmark{1}\authorrefmark{2}, 
Kai Jiang \authorrefmark{4},  
Ioannis Kymissis\authorrefmark{3}, 
Matei Ciocarlie\authorrefmark{2}% <-this % stops a space
\thanks{\authorrefmark{1} joint first authorship}%
\thanks{\authorrefmark{2}Dept. of Mechanical Engineering, \authorrefmark{3}Dept. of Electrical Engineering,\authorrefmark{4}Dept. of Computer Science}%
\thanks{Columbia University, New York, NY 10027, USA}%
\thanks{Email:~\texttt{\{aae2155,si2395,pp2511,kj2628,ik2174,}}%
\thanks{\texttt{mtc2103\}@columbia.edu}}%
\thanks{This work was supported in part by the NSF under awards PFI-232975 and CMMI-2037101.}
}

\usepackage{cite}

\usepackage{caption}
\usepackage{graphicx}
\usepackage[font=small]{caption}
\usepackage{amsmath}
\usepackage{amssymb}
\usepackage{xcolor}
\usepackage{hyperref}
\hypersetup{colorlinks,linkcolor={black},citecolor={green},urlcolor={gray}}

\usepackage{algorithm} % captioning
\usepackage[noend]{algpseudocode} % actual typesetting
\usepackage{lipsum}
\usepackage{subcaption}

\usepackage{booktabs}

\begin{document}
\maketitle
\thispagestyle{empty}
\pagestyle{empty}

%%%%%%%%%%%%%%%%%%%%%%%%%%%%%%%%%%%%%%%%%%%%%%%%%%%%%%%%%%%%%%%%%%%%%%%%%%%%%%%%
\begin{abstract}
In this paper, we introduce a sensor designed for integration into robotic fingers, where it can provide information on the displacements induced by external contacts. Our sensor uses LEDs to sense the displacement between two plates connected by a transparent elastomer; when a force is applied to the finger, the elastomer displaces and the LED signals change. We show that using LEDs as both light emitters and receivers in this context provides high sensitivity, allowing such an emitter and receiver pairs to detect very small displacements. We characterize the standalone performance of the sensor by testing the ability of a supervised learning model to predict complete force and torque data from its raw signals, and obtain a mean error between 0.05 and 0.07 N across the three directions of force applied to the finger. Our method allows for compact packaging (fitting at the base of a finger) with no amplification electronics, low cost manufacturing, easy integration into a complete hand, and high overload shear forces and bending torques, suggesting future applicability to complete manipulation tasks. 

\end{abstract}

%%%%%%%%%%%%%%%%%%%%%%%%%%%%%%%%%%%%%%%%%%%%%%%%%%%%%%%%%%%%%%%%%%%%%%%%%%%%%%%%

\section{Introduction}

Detecting the net forces or torques acting on a robotic finger is useful for dexterous manipulation \cite{ RobotHandSensorReview, shah2021design, fingertipforcesensing,
leslie2023optical, pirozzisensor2019}, with the potential to replace or complement tactile sensors. Tactile sensors convey significantly more information, such as the exact location of contact or a pressure map, from which net finger forces can also be extracted. However, tactile sensors offering full fingertip coverage are still not ubiquitous technology. In their absence, information on net forces and torques can still be useful in robot fingers engaged in manipulation tasks.

The most direct method for sensing such forces and torques is to add an off-the-shelf six-axis force and torque (F/T) sensor into the base of the finger. However, such F/T sensors are primarily developed with industrial applications in mind \cite{ft_industrial_intergation,Min2023,ft_design_principles}, leading to limitations when being used for dexterous manipulation. Manipulation often benefits from compliance \cite{chi2021compliance}, whereas most F/T sensors are optimized to be as stiff as possible, thus requiring precision engineering and leading to a high cost. Industrial applications also tend to favor a higher resolution than is needed for manipulation, but compromise by limiting the overload protection \cite{ft_industrial_intergation}. This leads to sensors that are easily damaged in the uncertain environments that robotics need to interact with. Finally, off-the-shelf F/T sensors can be difficult to package and integrate into the hand. Attempts to solve this problem by designing sensors to be placed within the fingertips \cite{ Fernandez2021Visiflex, Kim2020SixAxis,ft_mag1} limit the geometry of the finger and can also prevent the inclusion of other modalities within the finger.

In this paper, we present the design of a displacement sensor intended to provide information about the net forces acting on a robot finger in a manner that addresses the needs of robotic manipulation. First, such a sensor must be easy to manufacture using low-cost components and processes, so that it is affordable and practical to include multiple in a multi-fingered hand. Second, it must have the size and profile suited for the base of a fingertip, which will increase ease of integration into a hand. This also frees up the fingertip design for alternate sensors, or allows more flexibility in design of a finger. Third, some amount of compliance could be beneficial to manipulation, and thus there is no need to avoid compliance at all cost. Finally, the ideal sensor would have a Signal to Noise Ratio (SNR) that allows it to be useful for manipulation without sacrificing overload protection.

\begin{figure}[t!]
\setlength{\tabcolsep}{1.0mm}
\centering
    \includegraphics[width=\linewidth]{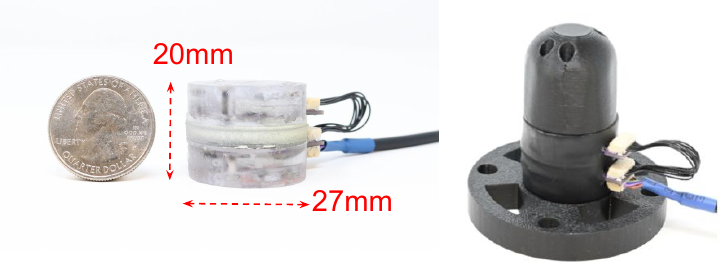}
\vspace{-0mm}
\caption{Our LED-based sensor, standalone (left) and mounted at the base of a robot finger (right). Our sensor uses LEDs mounted on two plates connected by a transparent elastomer to sense displacement between the plates that corresponds to the applied forces and torques. This method allows for a compact, fully integrated package at a low manufacturing cost.}
\label{fig:eyecandy}
\end{figure}

\begin{table*}[ht!]
  \centering
  \begin{tabular}{lccccccc}\toprule
\textbf{Modality} &\textbf{Cost} &\textbf{Size} &\parbox{20mm}{\centering \textbf{Min Forces\\and Torques}} &\parbox{20mm}{\centering \textbf{Max Forces\\and Torques}} &\textbf{Accuracy (R$^2$)} &\textbf{Bandwidth} \\\midrule
Barometer \cite{Guggenheim2017} &\$20 &\begin{tabular}{@{}c@{}} L : 75 mm \\W : 75mm \\H : 5 mm \end{tabular}&\begin{tabular}{@{}c@{}}Fx - $<$0.1N \\Fy - $<$0.1N \\Fz - $<$0.1N \\Mx - $<$0.1Nmm \\My - $<$0.1Nmm \\Mz - $<$0.1Nmm \end{tabular} &\begin{tabular}{@{}c@{}}Fx - 30 N \\Fy - 40 N \\Fz - 10 N (tension) \\Fz - 10 N (compression) \\Mx - 40 Nmm \\My - 40 Nmm \\Mz - 30 Nmm \end{tabular} &\begin{tabular}{@{}c@{}}Fx - 0.937 \\Fy - 0.909 \\Fz - 0.998 \\Mx - 0.984 \\My - 0.984 \\Mz - 0.991 \end{tabular} &50 Hz \\
\\[-2mm]
Camera \cite{Ouyang2020} &\$40 &\begin{tabular}{@{}c@{}} L : 35.7mm \\W : 22.5mm \\H : 51 mm\end{tabular} &\begin{tabular}{@{}c@{}}Fx - 0.1N \\Fy - 0.1N \\Fz - 0.1N \\Mx - 0.5Nmm \\My - 0.5Nmm \\Mz - 0.5Nmm \end{tabular}&\begin{tabular}{@{}c@{}}Fx - 65 N \\Fy - 18 N \\Fz - 0.1 N (tension) \\Fz - 80 N (compression) \\Mx - 80 Nmm \\My - 40 Nmm \\Mz - 4 Nmm mm\end{tabular}&\begin{tabular}{@{}c@{}}Fx - 0.991 \\Fy - 0.996 \\Fz - 0.875 \\Mx - 0.997 \\My - 0.997 \\Mz - 0.902 \end{tabular}&25 Hz \\
\\[-2mm]
Camera \cite{Fernandez2021Visiflex} &\$400 &\begin{tabular}{@{}c@{}}L : 35 mm \\W : 35 mm \\H : 35 mm\end{tabular} &Not reported &\begin{tabular}{@{}c@{}}Fx - 10 N \\Fy - 10 N \\Fz - 0.1 N (tension) \\Fz - 10 N (compression) \\Mx - Not reported \\My - Not reported \\Mz - Not reported \end{tabular}&\begin{tabular}{@{}c@{}}Fx - 0.993 \\Fy - 0.980 \\Fz - 0.988\\ Mx - 0.986 \\My - 0.980 \\Mz - 0.985 \end{tabular}&40 Hz \\
\\[-2mm]
Capacitance \cite{Kim2020SixAxis} &\$100 &\begin{tabular}{@{}c@{}}Rad : 14 mm \\H : 6.5 mm\end{tabular} &\begin{tabular}{@{}c@{}}Fx - 8.1 mN \\Fy - 10.2 mN  \\Fz - 5.4 mN \\Mx - .25Nmm \\My - .23Nmm \\Mz - .1Nmm \end{tabular} &\begin{tabular}{@{}c@{}}Fx - 30N \\Fy - 30N \\Fz - 30N (tension) \\Fz - 30N (compression) \\Mx -  0.30 N*m \\My -  0.30 N*m \\Mz -  0.30 N*m \end{tabular}&\begin{tabular}{@{}c@{}}absolute:\\Fx - 0.276 N \\Fy - 0.249 N  \\Fz - 0.471 N\\ Mx - 4.86 Nmm \\My - 5.13 Nmm \\Mz - 1.71 Nmm \end{tabular}&200  Hz \\
\bottomrule
\end{tabular}
  \caption{Comparison of novel 6-axis sensors developed to extract full contact information in the context of dexterous robot manipulation}
  \label{tab:1}
\end{table*}

To that end, we develop our sensor based on light transport. Our design, shown in Fig.~\ref{fig:eyecandy}, consists of two plates, connected by an elastomer layer that allows for 6 Degrees of  Freedom (DOF) between them. When mounted at the base of a robot finger, any contact force applied to the finger will cause a displacement between these plates. To measure this displacement, we use a network of LEDs on each plate as both light emitters and light receivers. As the plates shift relative to each other due to an externally applied forces, the signal measured by the receiver LEDs changes due to the relative movement and positioning of the emitter LEDs. We record these signals, and use them to extract information about the forces acting on the finger.

A key feature of our work is that we demonstrate that using LEDs as both emitters and receivers confers a high signal-to-noise ratio, significantly higher than using photodiodes. Combined with the small LED form factor, this allows us to sense very small displacements in a compact package. Using an off-the-shelf F/T sensor as ground truth, we show that our sensor has an observed resolution of .06 N and 2.6 N-mm in a compact, low-cost package that is compliant and easy to integrate in robot fingers as it requires no external amplification. To the best of our knowledge, this is the first time that LEDs have been integrated into a 6-DOF flexure, as both emitters and receivers, thereby allowing for full contact information in a compact and complaint robot finger. Our sensor is designed to be integrated at the base of the finger, which allows for sensing coverage of the entire finger and the ability to integrate additional sensors. We hope that this direction can facilitate truly ubiquitous sensing integrated in robot fingers, with future applications in manipulation.

\section{Related Work}

%Tactile sensors are devices that detect and measure contact forces and location, which improves robotic dexterity \cite{Chebotar2016TactileManipulation, Ward-Cherrier2018Tactile, Sundaralingam2019TactileDexterity, Schmidt2018TactileSLAM}. Notably, the GelSight360 \cite{GelSight360} allows for 360 degree coverage, but only can be used at the very tip of a finger. Any contact made anywhere but the fingertips cannot be sensed, and therefore the sensing coverage for the entire manipulator is very small. The same limitation applies to other tactile sensors, which significantly bottlenecks their utility in dexterous policies. Current tactile sensing technology does not allow for full sensing coverage for forces applied anywhere but the face of a phalanx. Compact F/T sensors provide an alternative to get full sensing coverage. 

%By integrating an F/T sensor in the phalanx, we can detect contact made anywhere past where the sensor is mounted - even contact on the back of the finger. However, commercially available miniature F/T sensors are very expensive. Several novel sensors, such as those listed in Table 1, give manipulators full sensing coverage, but do so at a much more affordable price. 

Many tactile sensors use LEDs due to their compact packaging and wide range of emission cones \cite{light_tactile_1,knu_sensor,gelsight,disco,gelslim}. The LEDs in these sensors are embedded in a  translucent elastomer, so that their emission cone shines on a receiving device. This receiver is often a photodiode \cite{knu_sensor,disco} or camera \cite{gelsight,gelslim,GelSight360}, but can also be an LED \cite{tactile_led_receiver}. Regardless of the receiving device, when a force is applied to the surface of these sensors, the elastomer deforms and the emitter and receiver will displace. Using either a data-driven approach or deriving an analytical relationship, the change in electrical signal from this deformation can then be mapped to the contact applied. These tactile sensors are not able to cover the full surface of a robot finger, and therefore do not provide full sensing coverage for all possible contacts. To detect the full contact, others have tried several modalities including strain-based transducers \cite{ft_strain1}, capacitance \cite{ft_capacticance1}, optical \cite{ft_opt1}, and magnetic \cite{ft_mag1}. Although most of these sensors are too large to integrate in a robot finger, there are a few novel F/T sensors that are built particularly for manipulation.

In Table 1, we organize other low-cost sensors developed specifically to address the needs of contact wrench detection in robot fingers. For example, Guggenheim et al. embedded barometers in a rubber flexure and placed them between two rigid plates \cite{Guggenheim2017}. These plates are relatively large with the length and width being 75mm. The size is constrained by the number of barometer modules needed to get 6-axis forces and torques. Camera-based sensors are more compact as they only need a single camera module, but require a completely translucent or hollow flexure. Some low-cost F/T sensors use small cameras to track the displacement of a marker fixed on some flexure. Ouyang et al. suspended a fiducial marker above a small camera module, using a compression spring as the flexure \cite{Ouyang2020}. However, this design requires a relatively tall structure at 51mm. The Visiflex, a camera-based F/T sensor developed by Fernandez et al., is much more compact \cite{Fernandez2021Visiflex}. However, it only tracks forces and torques applied directly on the surface of the sensor. Therefore, contact anywhere else on the manipulator will not be sensed. In contrast, Kim et al. developed a very compact and capable F/T sensor using cylindrical capacitors \cite{Kim2020SixAxis}. The overall package of the sensor is very small, especially compared to the other low-cost sensors listed in Table \ref{tab:1}. While force range is high, torque range is limiting, as is the absolute accuracy reported. 

Overall, we note that different sensors in Table \ref{tab:1} shine in different respects. In this context, the method introduced in this paper allows for compact size (27mm diameter and 20mm height), high bandwidth (2.5KHz) and good absolute accuracy, a low manufacturing cost (\$100 in small quantities), overload forces up to 110N, and overload torques up to 1.62 N*m. Overall, we believe each of these technologies holds promise for advancing manipulation, and different sensors will suit different use cases and application scenarios.

\section{Sensor Design}

\begin{figure}[t!]
    \centering
    \includegraphics[width=.8\linewidth]{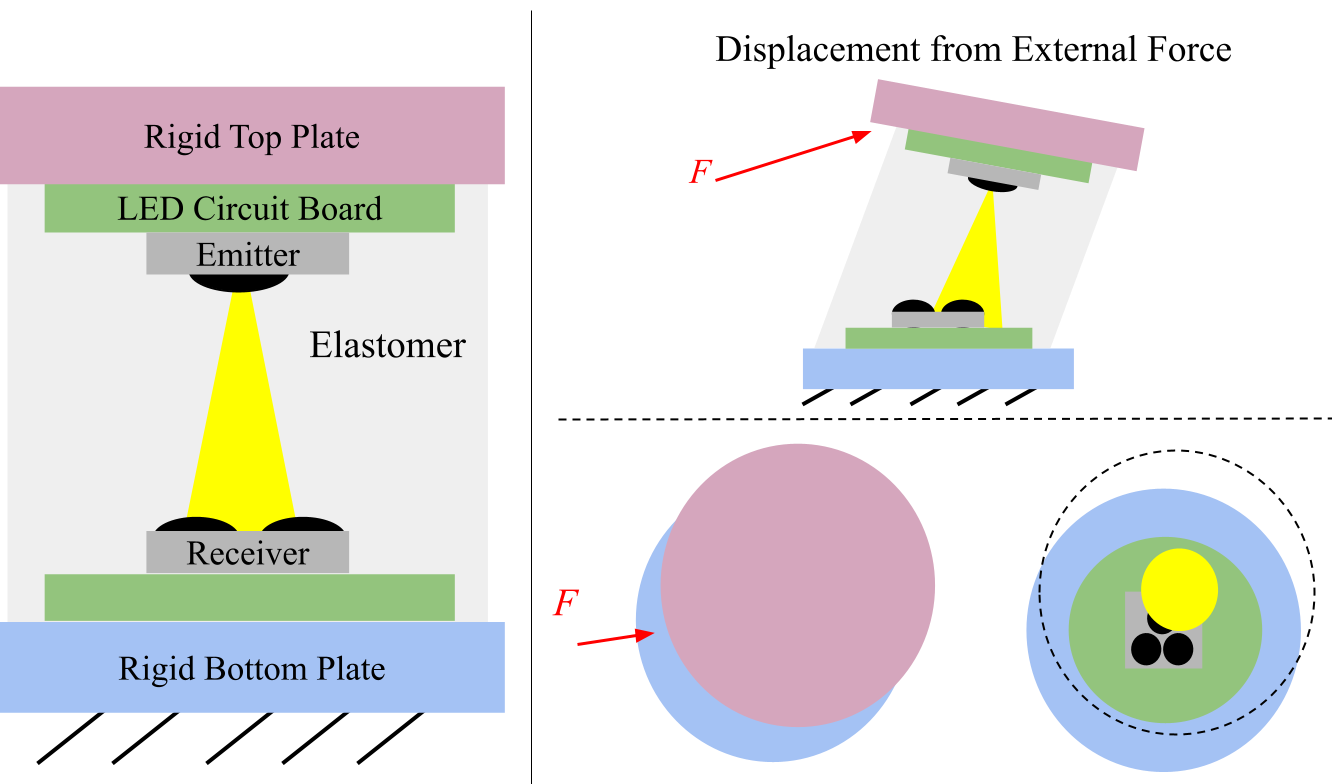}
    \caption{A diagram of our LED-to-LED sensor concept. We align two PCBs with LEDs directly above each other. We attach these circuit boards to rigid plates, we then suspend both of these plates in a transparent, elastic material. Assuming the bottom plate is fixed, a force applied to the top plate will displace the plate and change how to emitter LED shines on the receiving LEDs. We can then map this change to the contact wrench that induced the displacement.  }
    \label{fig:concept}
\end{figure}

\subsection{Concept}

The conceptual diagram of our sensor is shown in Fig. \ref{fig:concept}. It consists of two rigid plates connected by an elastomer layer that allows 6-DOF displacements. The electronics fixed to each rigid plate consist of a circuit boards with emitter LEDs and clusters of receiving LEDs. At the neutral position, the emitter LED is directly above the center of the receiving cluster. When an external force is applied, the plates will displace relative to each other. This displacement will result in a change in signal for the receiver LEDs due to the relative motion of the corresponding emitter LED. We can then map this change in signal back to the contact wrench applied. 

\subsection{Analysis of Single Emitter-Receiver Pair}

%Displacement transduction via light measurements offers three primary advantages. First, it provides a fast response time, as diode readings are gated by the ADC. Second, its components require minimal space, which aids with reducing our footprint. Third, we can select LEDs in the IR spectrum, making the sensor resistant to the majority of external environment light. Adding an opaque wrapping to the outside of our sensor can also prevent interference from external infrared light. 

Using LEDs as both a receiver and emitter gives us several advantages over the other modalities mentioned in Table \ref{tab:1}, as light can be sensed at very high frequencies with minimal electrical noise \cite{led_as_sensor}. There are several devices we could use as our receiver, but we opted for LEDs based on three factors: 1) LEDs have a smaller form factor than most photodiodes in the price range. 2) LEDs are optimized for operation in their wavelength range, making it easier to design features to minimize external interference. Adding an opaque wrapping to the outside of our sensor can further prevent interference from external light. 3) LEDs have a smaller viewing angle, which is useful for increasing sensitivity given our size constraints.

\begin{figure}[t!]
    \centering
    \includegraphics[width=\linewidth]{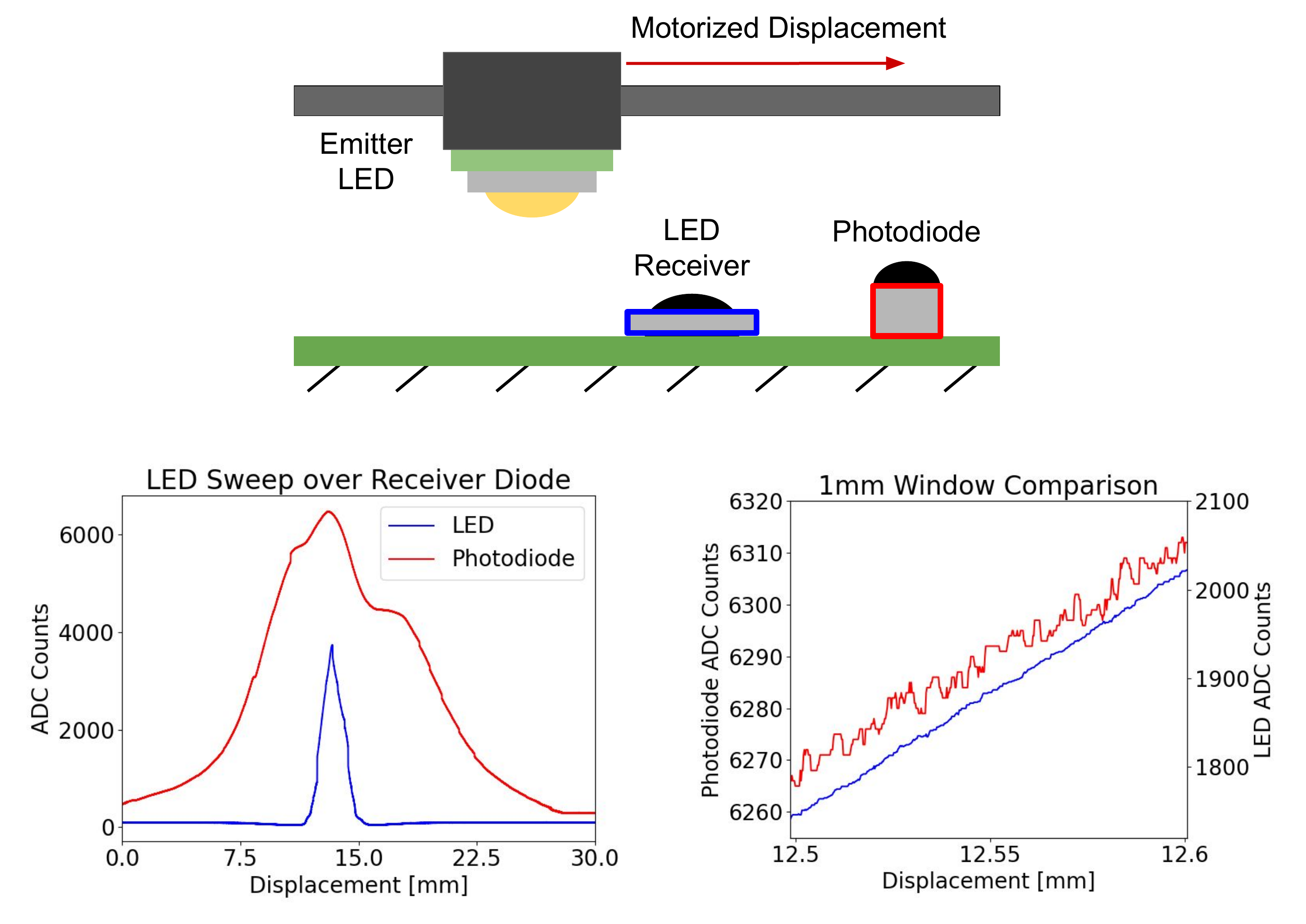}
    
    \caption{ To analyze single emitter-receiver pairs, we used a precise linear motor to displace an emitting LED while recording the response from both a receiving LED and a photodiode. \textbf{Left:} sharper cone response of the LED compared to the photodiode. \textbf{Right:} zooming in on response across a .1mm displacement, we see significantly higher SNR when using the LED as receiver versus the photodiode.}
    \label{fig:LED Sweep}
\end{figure}

%\begin{figure}
%    \centering
%    \includegraphics[width=\linewidth]{figs/two_axis_test_rig_2.pdf}
%    \caption{A two-axis, servo-driven testing rig we used to verify that an LED emitter-receiver pair will give us sufficient sensitivity to small displacements. The testing rig allows for both vertical and horizontal displacement. The vertical axis is driven by lead screws actuated by a Ax-12A. The horizontal axis is driven by a lead-screw and actuated by an XM-430A. This testing rig allows us to measure signal changes at .1mm displacements.}
%    \label{fig:testrig}
%\end{figure}

We first verified the validity of LEDs as both emitters and receivers, and compared to the alternative of using photodiodes on the receiving end. To achieve this, we measured the response to relative horizontal displacements of a single emitter-received pair separated by an air gap, where the receiver consisted of either an LED or a photodiode. The results, plotting receiver signal vs. relative horizontal displacement of the emitter, are shown in Fig. \ref{fig:LED Sweep}. We observe that the LED receiver exhibits much higher sensitivity compared to its photodiode counterpart. In fact, the LED receiver can identify displacements as small as 0.01 mm, whereas the corresponding signal changes for a photodiode receiver are below the noise level. Based on this result, we proceed with LED-LED pairs for the overall sensor design.

\begin{figure}[t]
    \centering
    \includegraphics[width=\linewidth]{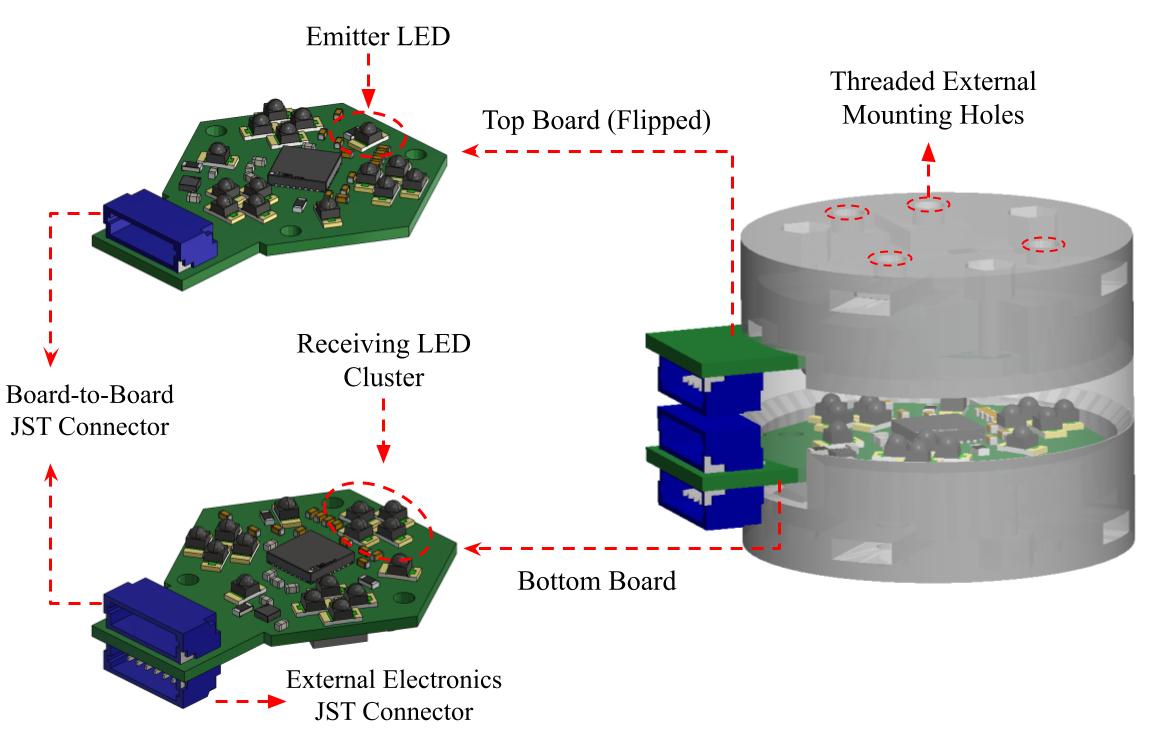}
    \caption{ Our complete sensor is composed of two custom PCBs that both contain LED emitters and clusters of LED receivers. The PCBs are fastened to rigid, 3-D printed plates. The plates are then fixed above each other in a mold that we then fill with polydimethylsiloxane (PDMS). We use a single JST connector to communicate between the top and bottom board, and an additional JST connector on the bottom board to communicate with external electronics. The rigid plates are fitted with threaded holes to mount external hardware. Our sensor is 27 mm in diameter, 20mm in height, and weighs 16.4g. }
    \label{fig:complete-design}
\end{figure}

\subsection{Complete Sensor Design}

With the core concept of using identical LEDs as emitters and receivers verified, we implemented a complete design for the sensor. As we wanted to mount this at the base of a finger, we constrained our design to a one inch (25.4 mm) diameter. The goal was to maximize the amount of data signals within the space. Based on our LED analysis, we used a compact, infrared LED (SFH 4056) with an emission cone of $20 ^\circ$. The primary size constraint became the operational amplifiers, which meant we could support a maximum of 12 signals per board. As shown in the left of Fig. \ref{fig:complete-design}, we divided the signals into clusters of four receivers, so that each board had three clusters, for a total of 24 signals broken across 12 clusters. We then suspended the PCB boards above each other and attached them with an elastic flexure. 

We fastened the bottom and top PCB boards, as shown in Fig. \ref{fig:complete-design}, each to a rigid, 3-D printed plate. The plates each have three M2 inserts used to mount to external hardware. We used these inserts to attach these rigid plates to a mold that we then filled with an uncured mixture of polydimethylsiloxane (PDMS). We cured the filled mold at $70 ^\circ C$ for 8 hours, which gave us the sensor shown in Fig. \ref{fig:complete-design}. The overall dimensions of the sensor are 20.82mm in height and 27mm in diameter, just above our target diameter. We then covered the sensor with a thin, opaque rubber shrink-wrap tube. This cover inhibits interference from external light sources and adds to the sensor robustness. We cured the PDMS at a 10:1 ratio of base to curing agent, as we observed this best achieved the desired sensor stiffness and surface finish. Moreover, this ratio produced the most consistent behavior across multiple sensors. 

To increase PDMS adherence to the plates, we treated the PCB boards with silicone primer to promote strong adhesion between the PDMS and PCB faces. We also added a lip to the rigid plates that extends past the PCB surface to help the PDMS from shearing off the PCB. The rubber wrapping also increases the stiffness of our sensor and adds robustness under large displacements, especially in shear. 

Our electronics also need to take into account concerns with integration in a robot hand. To that end, we want to minimize the number of signals transmitted between the board and the controller. Using an on-board ADC provides four advantages: 1) It allows for higher quality signals than utilizing the built-in ADCs on the micro-controller. 2) It allows us to use SPI protocol with only 7 wires. This also enables chaining of the top plate and bottom plate of the sensor using the SPI protocol. 3) It allows us to only transmit digital signals across the cables, minimizing transmission noise. 4) We maintain higher response time by avoiding external multiplexing.

\begin{figure}[t]
    \centering
    \includegraphics[width=\linewidth]{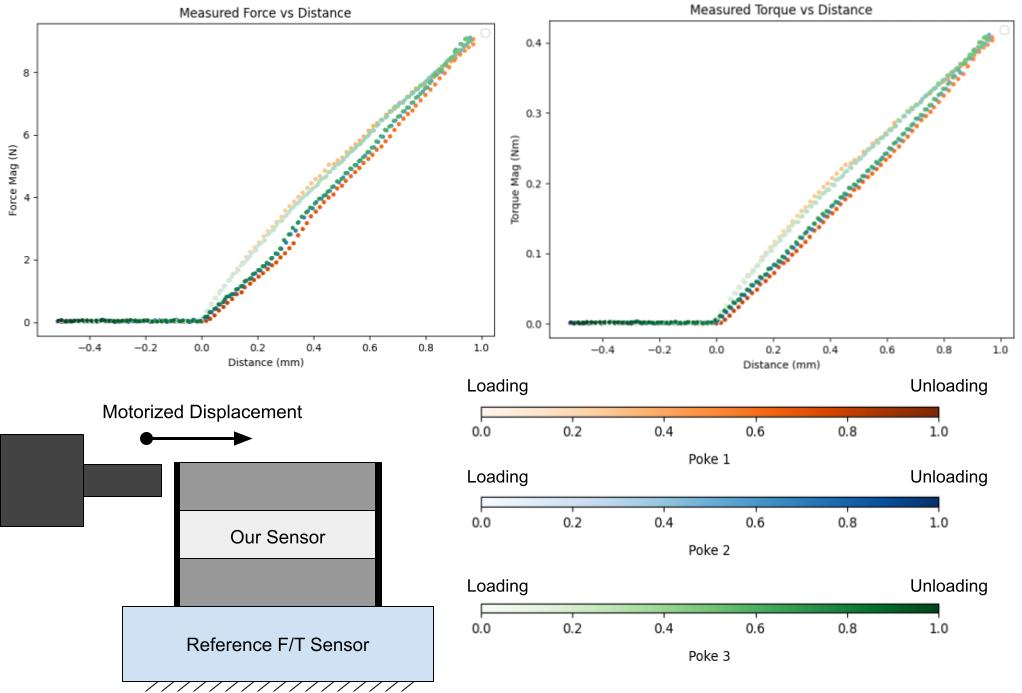}
    \caption{Mechanical hysteresis testing over three consecutive loading and unloading cycles. We fixed our sensor on a reference F/T sensor and used a linear probe to displace the top plate by 1 mm.}
    \label{fig:hysterisis}
\end{figure}

\begin{figure} [t]
    \centering
    \includegraphics[width=\linewidth]{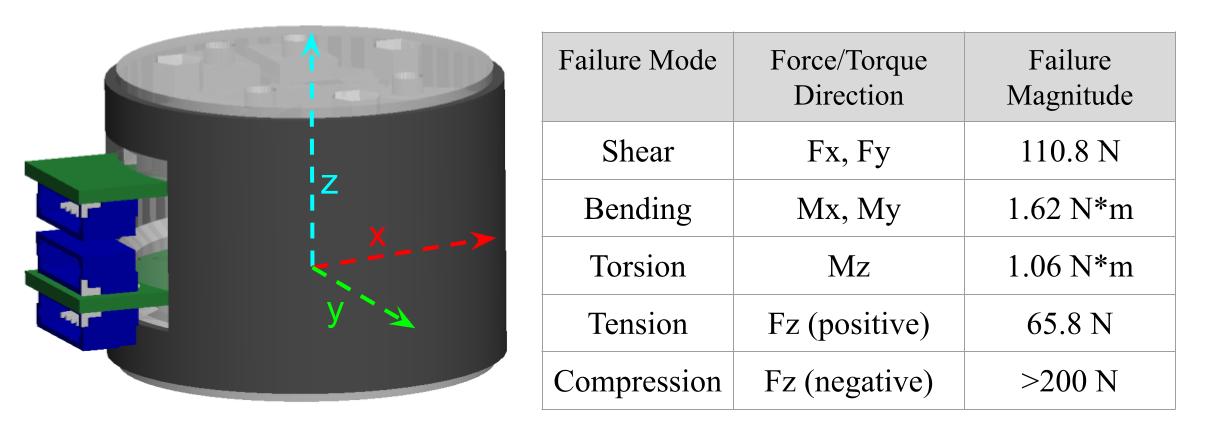} 
    \caption{Mechanical overload forces and torques for our sensor.  We built five sensors and then used an ADMET Instron machine to apply a horizontal force until the sensor failed. We tested our sensor in five major failure modes - shear, bending, torsion, tension, compression. All tests were done with the opaque, shrink wrapped rubber cover on.}
    \label{fig:overload}
\end{figure}

\begin{figure*}
    \centering
    \includegraphics[width=1.0\linewidth]{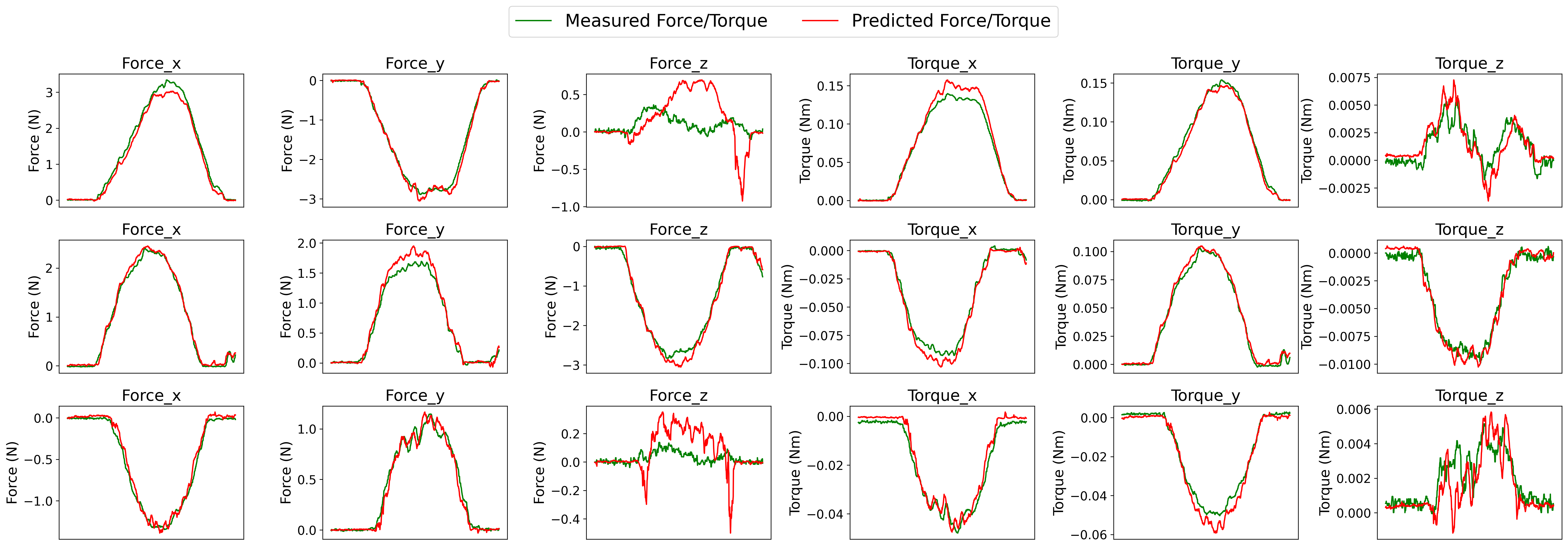}
    % \includegraphics[width=0.16\linewidth]{figs/Fx_good_1N5.png}
    % \includegraphics[width=0.16\linewidth]{figs/Fy_good_1N5.png}
    % \includegraphics[width=0.16\linewidth]{figs/Fz_good_1N5.png}
    % \includegraphics[width=0.16\linewidth]{figs/Tx_good_1N5.png}
    % \includegraphics[width=0.16\linewidth]{figs/Ty_good_1N5.png}
    % \includegraphics[width=0.16\linewidth]{figs/Tz_good_1N5.png}
    % \includegraphics[width=0.16\linewidth]{figs/Fx_bad_s1N5.png}
    % \includegraphics[width=0.16\linewidth]{figs/Fy_bad_s1N5.png}
    % \includegraphics[width=0.16\linewidth]{figs/Fz_bad_s1N5.png}
    % \includegraphics[width=0.16\linewidth]{figs/Tx_bad_s1N5.png}
    % \includegraphics[width=0.16\linewidth]{figs/Ty_bad_s1N5.png}
    % \includegraphics[width=0.16\linewidth]{figs/Tz_bad_s1N5.png}
    % \caption{Predicted vs ground truth Force and Torques across all 6 DoFs. Results presented from a test set with low accuracy on force but high on torque and a test set with high accuracy on force but low accuracy on torque for comparison. \textbf{Top}: Results from a contact with mean force error of .547N and torque error of 1.23Nmm. The prediction follows the shape tightly for Fx and Fy, but struggles at points with the magnitude. The torques meanwhile are much more accurate, following both shape and magnitude. \textbf{Bottom}: Force error of .0494N and torque error of 2.10Nmm. This prediction consistnetly undervalues the magnitude of both force and torques.}
    \caption{Predicted vs. ground truth forces and torques across all 6 DoFs. Each row represents a test poke on our sensor. The contact prediction model demonstrates strong performance for forces and torques along/about the x and y-axes but struggles particularly with force prediction along the z-axis.}
    \label{fig:pokes}
\end{figure*}

\section{Mechanical Characterization}
\label{sec:mechanical}

A core design objective of this sensor is robustness against large displacements induced by high contact forces. We tested our sensor to failure in five failure modes - shear, bending, torsion, tension, and compression. We note that, since our sensor is intended to be mounted at the base of a finger, as shown in Fig. \ref{fig:eyecandy}, contact forces will mostly cause shear and bending of our sensor. Additionally, all tests on our sensor were done with a shrink wrapped neoprene cover, as shown on the finger in Fig. \ref{fig:eyecandy}. 

Results are shown in Fig. \ref{fig:overload}.  In particular, we note that our sensor withstands shear forces of up to 110 N and bending torques of up to 1.62 N*m. The overload forces for torsion and compression are also high. Our sensor is weakest in tension, which is not a major concern as a robot finger would likely not experience large tension forces. 

PDMS is a hysteric material, so we expect our 6-DOF flexure to display some time-dependent non-linearity \cite{light_tactile_1,SCHNEIDER200995}. In Fig. \ref{fig:hysterisis}, we plot the results of cyclic loading and unloading tests for our sensor. We used a linear probe to horizontally displace the top plate of our sensor while it was fixed to a reference, off-the-shelf F/T sensor. Displacing the top plate by 1 mm three consecutive times, we observe hysteresis that is relatively consistent over the cycles. While hysteresis is a common effect for sensors that use PDMS as an elastic flexure \cite{SCHNEIDER200995}, it is also problematic for mapping the electrical signals to the force applied, and often displays behavior that is unique to each sensor. To design a more robust and consistent sensor, we intend to investigate a variety of different flexures, including other clear polymers as well as rigid alternatives such as metal or plastic flexures. Additionally, future work can aim to address hysteresis from a software perspective by adding time-history information to the model processing the raw data. 

\section{Performance}

Ultimately, we believe that the true value of a sensor designed for robot fingers lies in the performance improvements it enables for complete manipulation. If manipulation is performed via learned policies, such policies are likely to use raw sensor data as input (in our case, raw readings from the receiver LEDs), decreasing the importance of any intermediate representations. However, training such a policy exceeds the scope of this paper, which focuses on the design of the sensor itself. Therefore, in this section, we aim to quantify the performance of our design as a standalone sensor. 

Since we are likely to train policies directly on raw signals produced by contacts on this sensor, we did not formulate an analytical model relating the change in signal to the displacement of the PCBs and the force applied. Instead, in order to quantify the standalone performance of our sensor, we test the ability to infer ground truth wrenches applied to the robotic finger from the raw sensor data. To that end, we mount our sensor on a commercial ATI SI-130-10 to obtain ground truth force and torque measurements. We then trained a simple machine learning model to predict applied force and torque based on our sensor data. We describe the experimental approach and results in this section.

\subsection{Data Collection}

To collect the dataset, we mounted a 3D-printed finger-like shape on top of our sensor and made contact with it from all directions along its circumference (Figure~\ref{fig:eyecandy}) while it was mounted on a reference FT sensor. Each data file contains three contact events recorded within a 15-second span. To ensure force and torque were decoupled, we applied a range of forces at different heights. The dataset primarily captures forces between 0–5 N and torques between 0–0.2 Nm along each axis, with a maximum force magnitude limited to under 10 N.

We selected ten contact locations, distributed evenly around the circumference of the finger. Contacts were applied by pressing against the printed finger with our fingertip for two and five seconds. For the training set, we collected data for 15 different contacts at each location. These contacts were grouped into sets of three, each recorded at different heights—top, middle, and bottom of the finger. Each set focused on a specific force range to ensure comprehensive coverage. The test sets are collected with the same procedure, but included only one contact per surface for each contact position. The test set was randomly split across the height groupings to be representative of the different torque-force pairings.

During data collection, the sensor operated at a sampling rate slightly above 2.5 kHz. However, data was down-sampled and recorded at 250 Hz due to the practical requirements of robotic manipulation tasks.

\subsection{Data pre-processing}

Prior to training, we applied a series of pre-processing techniques to improve data quality and model performance. To reduce high-frequency noise, we used an 11-point median filter on the raw sensor readings and a 5-point median filter on the ground truth data. Given our data collection rate of 250 Hz, this filtering process introduces a delay of less than 0.1 seconds, which is negligible for our application.

To distinguish between contact and non-contact states, we thresholded all data points with a total force magnitude below 0.1 N as “no contact.” However, to prevent the dataset from being overwhelmingly dominated by these points, we downsampled them to ensure they constitute only 10\% of the training data.

Additionally, instead of training directly on absolute raw signals, we extracted features based on their relative changes from a no-contact baseline. This baseline was defined as the mean of the first 50 data points in each recorded dataset. To further stabilize training and maintain numerical consistency, we normalized all features and labels prior to training. These pre-processing steps help the model focus on signal variations rather than absolute values, enhancing generalization.

\subsection{Learning Algorithms}
For all experiments on the sensor, we employ an updated ResNet-based architecture inspired by He et al. \cite{he2016deep}. Our model consists of a shared feature extraction backbone followed by six independent task-specific heads, one for each of the six DOFs in force and torque prediction. 

% \begin{figure}
%     \centering
%     \includegraphics[width=\linewidth]{figs/MultiTask-Resnet.png} 
%     \caption{Overview of our network architecture, consisting of a shared feature extraction backbone followed by six independent heads for force and torque prediction. }
%     \label{fig:multitask_resnet}
% \end{figure}

\begin{figure}
    \centering
    \includegraphics[width=0.49\linewidth]{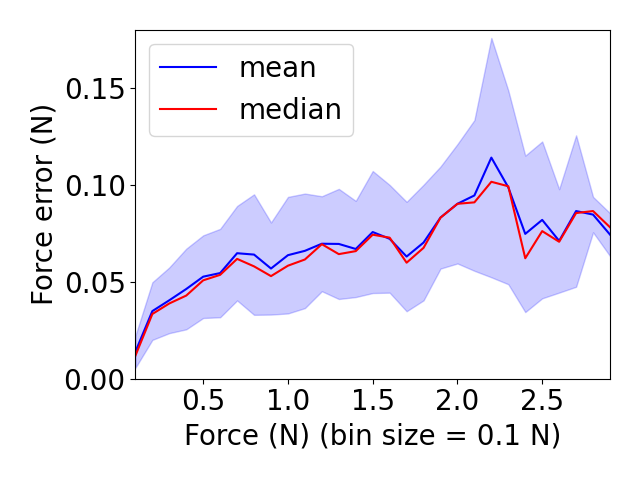}
    \includegraphics[width=0.49\linewidth]{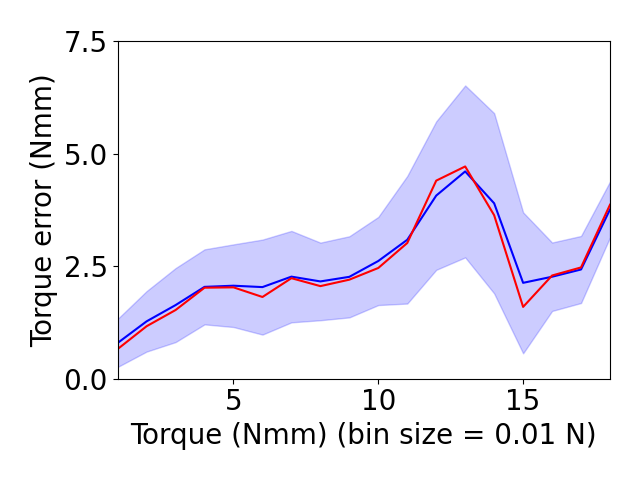}
    \caption{Graphs of error plotted against magnitude of the signal. \textbf{Left:} Force. \textbf{Right:} Torque. The plots demonstrate that for both force and torque, the magnitude of error scales slower than  the  magnitude of the force. What starts at a 20\% mean error rapidly becomes a 5-10\% mean error after 1N of force.}
    \label{fig:7}
\end{figure}

\begin{table}
    \centering
    \begin{tabular}{cccc}
        \bf{Variable} & \bf{Mean Abs. Error} & \bf{Std. Dev.} & \bf{R$^2$}\\
        Force x (N) &  0.06 & 0.07 & .990 \\
        Force y (N) & 0.05 & 0.07 & .993 \\
        Force z (N) & 0.07 & 0.09 & .560 \\
        Force magnitude (N) & 0.06 & 0.08 & .978 \\
        Torque x (Nmm) & 2.62 & 3.45 & .995 \\
        Torque y (Nmm) & 2.68 & 3.41 & .991 \\
        Torque z (Nmm) & 0.64 & 0.73 & .923 \\
        Torque magnitude (Nmm) &  2.90 & 3.85  & .986\\
    \end{tabular}
    \caption{Accuracy Reporting for the F/T Model. X, Y, and magnitude results are quite promising, but Z still requires further data collection and training before conclusions can be drawn. Despite the absolute error being in the same threshold, the magnitude of Z forces is much smaller leading to a higher percentage error. This is most evident in the R2 results which shows the much higher correlation for x and  y than for z}
    \label{tab:my_label}
\end{table}

The key component of our model is the residual block with 1D convolutional layers. Each residual block integrates 1D convolutional layers, skip connections and ReLU activation functions to enhance feature extraction while mitigating vanishing gradients.

The input to the model is a 24-channel time-series sequence, corresponding to the preprocessed sensor signals. A 1D convolutional layer with an enlarged kernel is applied before the residual blocks to capture global patterns in the sensor signals. This is followed by a shared backbone, which is a 1D ResNet specifically designed for sequential sensor data. The backbone consists of multiple residual blocks that efficiently capture spatial and temporal dependencies. 

After the shared backbone, the model branches into six independent prediction heads, each dedicated to estimating a specific force or torque component. Each head consists of additional residual layers, a global average pooling layer, and a fully connected output layer that generates a single scalar value. This multi-task learning framework enables specialized feature extraction while leveraging shared representations, thereby enhancing generalization across different force and torque components. 

The model is trained using a mean squared error (MSE) loss function, applied separately to each output head. Training is conducted for 100 epochs using the Adam optimizer with a batch size of 2048 and an initial learning rate of 1e-3, which is adjusted dynamically using a LambdaLR scheduler. 

\subsection{Results and Discussion}

The overall results of our sensor's contact prediction are presented in Table~\ref{tab:my_label} and Figure~\ref{fig:7}. In particular, Table~\ref{tab:my_label} details the performance for each force and torque direction, as well as the overall force and torque magnitudes. Figure~\ref{fig:7} visualizes the mean absolute error as a function of applied force and torque. In addition,  Figure~\ref{fig:pokes} provides a detailed performance analysis, showcasing all six force/torque dimensions across three exemplar contacts, each differing in the contact positions. We note again that the test data consisted of contacts never seen in training. Moreover, as we are leveraging LEDs as receivers, our model can operate with inputs streaming in at 250 Hz. 

We observe that the sensor provides accurate measurements in both force and torque for the x and y axes, but less so for the z axis. Predictions are accurate in terms of overall force magnitude, but can occasionally suffer from inaccuracy at higher forces in individual DOFs.  The model was evaluated primarily on the 0-5 N range of forces, but experimentation showed that training on data that extended beyond this range proved useful for the models learning. 

We believe that one likely source of error is hysteresis introduced by the PDMS and quantified in Sec.~\ref{sec:mechanical}. This is particularly prevalent with larger forces, where signals will take some time to settle back to their baselines, which introduces noise to forces done before the resettling, and also introduces noise into the zero-force data.

\section{Conclusion and Future Work}

In this paper, we introduce a low-cost, compliant sensor designed to be easily integrated in a robot finger and provide information on the displacement induced by contact with the world. Our method relies on sensing the displacement between two rigid plates, each mounted with LED emitters and receivers. We analyzed such an LED-LED pair in isolation, and found it to be highly sensitive to relative displacements of the emitter and the receiver. We thus designed a sensor with six LEDs acting as emitters and 24 LEDs acting as receivers, distributed between the top and bottom plates.

To characterize the information contained in the raw signals from our sensor, we tested whether it is possible to reconstruct overall finger wrench information, using a commercial force/torque sensor as ground truth. Our results showed that a simple supervised learning model trained on data from the 24 receiver LEDs can learn to map changes in these signals due to displacement of the flexure to external forces and torques applied between the two plates. The sensor output was particularly accurate for force/torques along/around the x and y axes, with additional work needed to improve sensitivity for the z axis. 

Although our sensor does not outperform the accuracy or precision of commodity F/T sensors, we believe that it offers useful information - especially along/about the axes most prevalent for manipulation - in a compact, robust and low-cost package. Modern learning-based manipulation methods often operate directly on raw, even uncalibrated sensor data  \cite{bhirangi2024anyskinplugandplayskinsensing, pattabiraman2024learningprecisecontactrichmanipulation}, thus converting to an intermediate representation such as the one used here for sensor characterization may be unnecessary in the future. Considering the strong SNR of the raw signals of our sensor, we believe that it can be used in a similar manner for policy learning for manipulation.

Future work of the sensor falls under two categories: further improvement of the sensor, and integration within functional robot hands. Sensor improvement in turn will have mechanical, electronic, and learning components. The focus along all these domains will be to make a more robust, consistent, and easy to manufacture device. Although we would like to eliminate hysteresis, ultimately some level of time dependent non-linearity will have to be accepted. Therefore, in making decisions different flexure designs in the future, we will prioritize robustness, ease of manufacture, and consistency. Electronically, we will continue to look for sources of noise in LED signals. We also want to minimize the form factor as much as possible, so further experiments would include use of fewer LEDs in each cluster (or potentially even fewer clusters). Learning wise, we will further tune the model as we gather more data for z axis and explore alternate architectures with both time series and single-snapshot models.

\begin{figure}[t]
    \centering
    \includegraphics[width=0.45\linewidth]{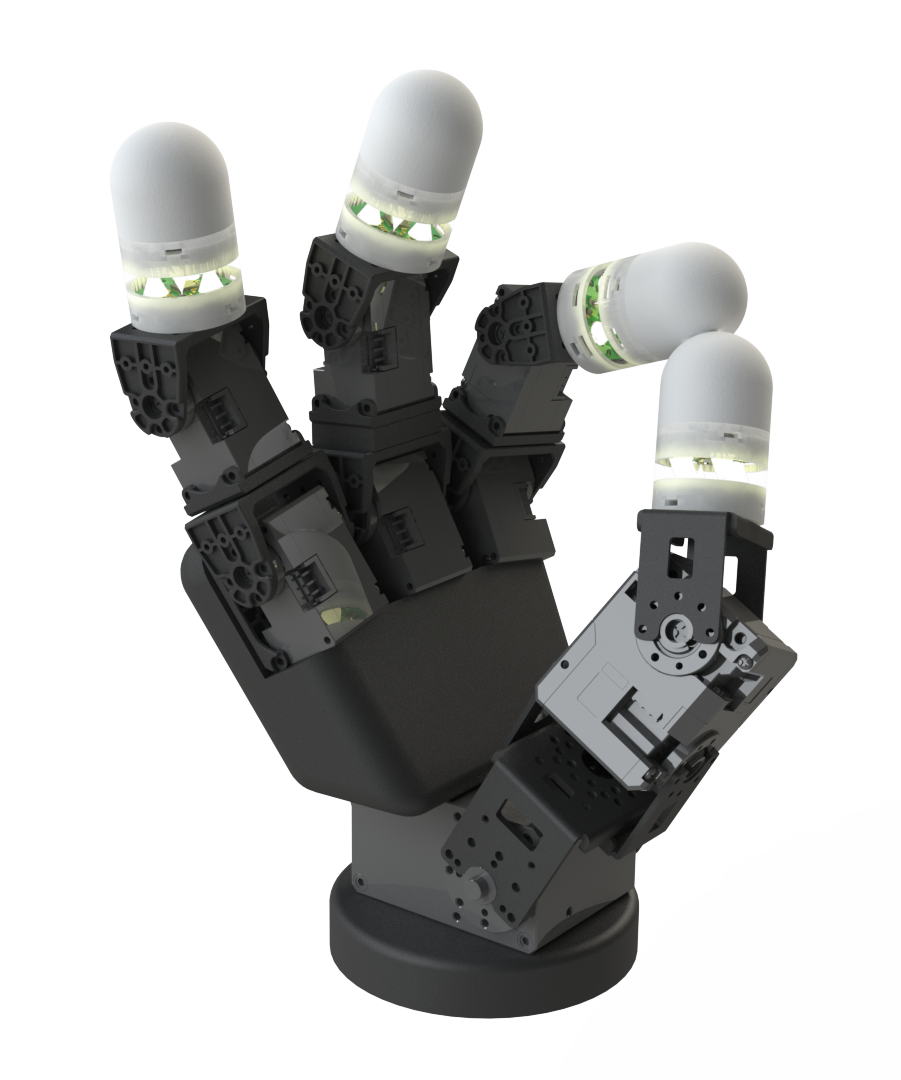}
    \includegraphics[width=0.45\linewidth]{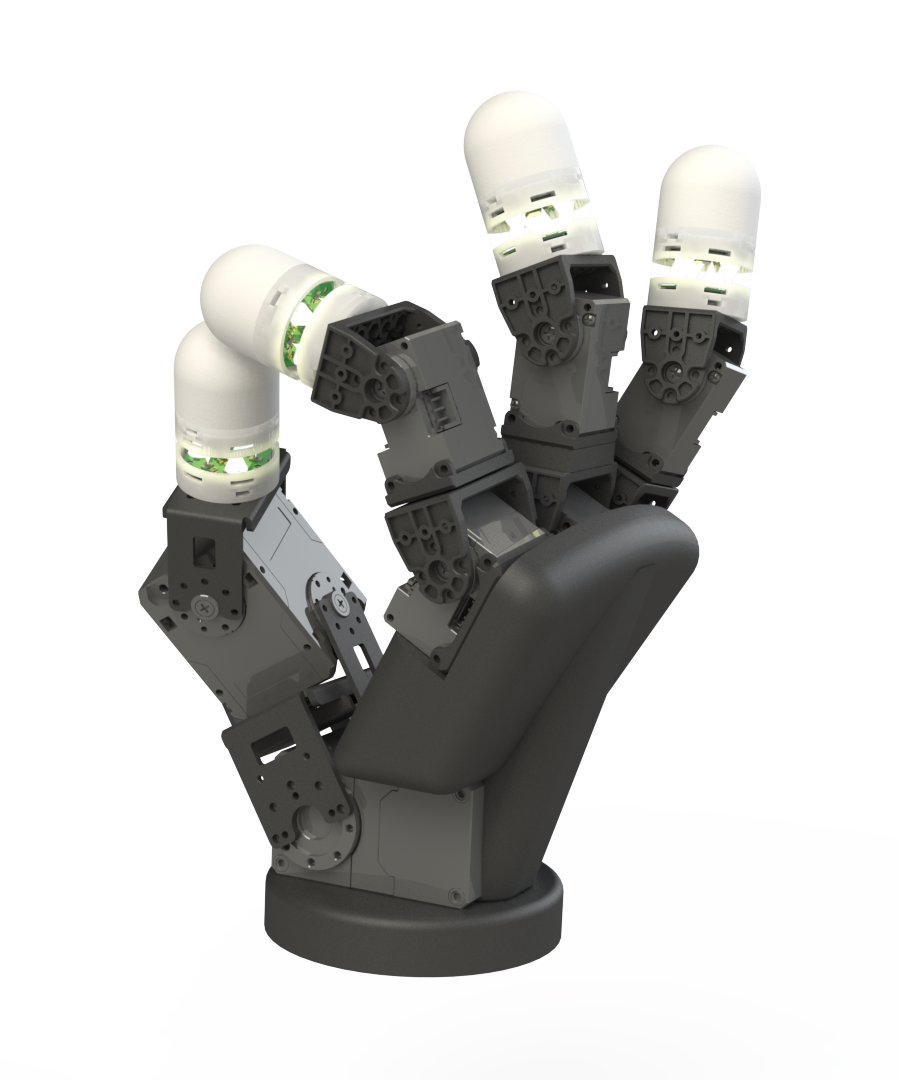}
    \caption{A render of an anthropomorphic hand design with the F/T sensors proposed in this paper mounted at the base of every finger.}
    \label{fig:hands}
\end{figure}

We look to take advantage of the features of the sensor which make it particularly attractive for integration in robot hands. In particular, the form factor and ease of integration (the sensor directly connects to a micro-controller board with no additional amplification electronics) are highly appealing. Figure~\ref{fig:hands} shows a possible design of an anthropomorphic hand with one of our sensors at the base of the distal phalanx of each finger. Such integration will also allow us to test the hypothesis that slight compliance in the sensing element can prove beneficial to dexterous manipulation.

As motor learning for dexterous manipulation makes continuous advances, we believe that sensing will prove key for future performance of robot hands. Having a multitude of sensors at our disposal, each with various strengths, will increase the ability of combining the right sensor with the right motor learning method to achieve novel capabilities. \textit{In particular, sensors that provide information on overall contact forces should be ubiquitous, commodity technology, available in any lab by the dozen, ready to be quickly integrated in any hand design, as opposed to expensive, fragile pieces connecting to complex external electronics and thus used rarely, if at all.} We believe that the design presented in this paper represents a step in this direction.

\bibliographystyle{IEEEtran}
\bibliography{references}

\addtolength{\textheight}{0cm}   % This command serves to balance the column lengths
%\clearpage
% \printbibliography

\end{document}